\documentclass[letterpaper, 10 pt, conference]{URL-ieeeconf} % Ensure URL-ieeeconf.cls is in the same directory
\IEEEoverridecommandlockouts    % This command is only needed if
\usepackage[T1]{fontenc}
\usepackage{cite}
\usepackage{amssymb,amsfonts}
\makeatletter
\let\NAT@parse\undefined
\makeatother
\usepackage{hyperref}
\hypersetup{
    colorlinks=false,
    linkcolor=blue,
    filecolor=magenta,      
    urlcolor=cyan,
    pdftitle={Overleaf Example},
    pdfpagemode=FullScreen,
    }
\usepackage{algorithmic}
\usepackage{graphicx} % Required for the \rotatebox command
\usepackage{textcomp}
\usepackage{mathtools}
\usepackage{changes}
\usepackage{subcaption}
\usepackage{algorithm}
\PassOptionsToPackage{table}{xcolor}
\usepackage{xcolor}
\usepackage{color, soul}
\usepackage{amsmath}
\usepackage{stmaryrd}
\usepackage{booktabs}
\usepackage{multirow}
\usepackage{xstring}
\usepackage{hhline}
\usepackage{kotex}
\usepackage{comment}
\usepackage{tikz}
\usepackage{cleveref}
\usepackage{adjustbox} % To adjust the table size if needed
\usepackage{makecell} % To allow line breaks in table cells
\usepackage{array} % To manually adjust column width
\usepackage{caption} % For better control of caption width and position
\usepackage[table]{xcolor}
\usepackage{colortbl}

\usepackage{bm}         % For bold math symbols (e.g., \mathbf)
\usepackage{xspace}     % For the \xspace command to avoid issues with space after macros
\floatname{algorithm}{Procedure}

\definecolor{rvc}{RGB}{0, 0, 255}
\definecolor{suyong}{RGB}{0, 255, 0}
\definecolor{comment}{RGB}{0, 0, 0}
\definecolor{cv2}{RGB}{0, 0, 0}
\definecolor{qw}{RGB}{0, 0, 0}
\definecolor{qwr}{RGB}{0, 0, 0}
\definecolor{qwe}{RGB}{0, 0, 0}

\sethlcolor{lightgray}

\makeatletter
\DeclareRobustCommand{\iscircle}{\mathord{\mathpalette\is@circle\relax}}
\newcommand\is@circle[2]{%
  \begingroup
  \sbox\z@{\raisebox{\depth}{$\m@th#1\bigcirc$}}%
  \sbox\tw@{$#1\square$}%
  \resizebox{!}{\ht\tw@}{\usebox{\z@}}%
  \endgroup
}
\makeatother

\IEEEoverridecommandlockouts                              % This command is only needed if 
\title{\LARGE \bf MambaGlue: Fast and Robust Local Feature Matching With Mamba}

\author{Kihwan Ryoo$^{1}$, Hyungtae Lim$^{2}$, and Hyun Myung$^{1*}$ % arXiv
  \thanks{$^*$Corresponding author: Hyun Myung}
  \thanks{$^{1}$Kihwan Ryoo and Hyun Myung are with the School of Electrical Engineering, KAIST (Korea Advanced Institute of Science and Technology), Daejeon, 34141, Republic of Korea, Email: {\tt\scriptsize \{rkh137, hmyung\}@kaist.ac.kr} \hfill \break
  \indent $^{2}$Hyungtae Lim is with the Laboratory for Information \& Decision Systems~(LIDS), Massachusetts Institute of Technology, Cambridge, MA 02139, USA, Email: {\tt\scriptsize \{shapelim\}@mit.edu} \hfill
  }
}

\begin{document}
\maketitle
\thispagestyle{empty}
\pagestyle{empty}

%%%%%%%%%%%%%%%%%%%%%%%%%%%%%%%%%%%%%%%%%%%%%%%%%%%%%%%%%%%%%%%%%%%%%%%%%%%%%%%%
\begin{abstract}
In recent years, robust matching methods using deep learning-based approaches have been actively studied and improved in computer vision tasks. 
However, there remains a persistent demand for both robust and fast matching techniques.
To address this, we propose a novel Mamba-based local feature matching approach, called \textit{MambaGlue}, where Mamba is an emerging state-of-the-art architecture rapidly gaining recognition for its superior speed in both training and inference, and promising performance compared with Transformer architectures. 
In particular, we propose two modules: a)~MambaAttention mixer to simultaneously and selectively understand the local and global context through the Mamba-based self-attention structure and b)~deep confidence score regressor, which is a multi-layer perceptron~(MLP)-based architecture that evaluates a score indicating how confidently matching predictions correspond to the ground-truth correspondences. 
Consequently, our MambaGlue achieves a balance between robustness and efficiency in real-world applications.
As verified on various public datasets, we demonstrate that our MambaGlue yields a substantial performance improvement over baseline approaches while maintaining fast inference speed.
Our code will be available on \href{https://github.com/url-kaist/MambaGlue}{\texttt{https://github.com/url-kaist/MambaGlue}}.
\end{abstract}

%%%%%%%%%%%%%%%%%%%%%%%%%%%%%%%%%%%%%%%%%%%%%%%%%%%%%%%%%%%%%%%%%%%%%%%%%%%%%%%%
\section{Introduction}
\label{sec:intro}

%%%%%%%%%%%%%%%%%%%
%% WHY: 
% First, answer the WHY question: Why is that relevant? Why should I be
% motivated to read the paper? Why should I care? (1 paragraph, 2-5 sentences)

%%%%%%%%%%%%%%%%%%%
%% WHICH PROBLEM
% Second, explain WHICH problem you are solving/address to solve.

%%%%%%%%%%%%%%%%%%%
%% HOW & WHAT
% Third, explain briefly how one can address the problem in general and mention 
% briefly what others/we before have done. Prepare the reader for your contribution 
% that comes in the next section (and not here!).

% Link to figure somewhere
% See \figref{fig:motivation} for an example.

Feature matching is a crucial component of various geometric computer vision tasks that involve estimating correspondences between points across images of a 3D map, including visual localization~\cite{sarlin2019coarse,sarlin2021back}, simultaneous localization and mapping (SLAM)~\cite{qin2018vins,mur2015orb,lim2022uv}, structure-from-motion~(SfM)~\cite{agarwal2011building,schonberger2016structure}, and more. Typically, these vision techniques involve matching local features detected in image pairs using descriptor vectors that encode their visual appearance. To achieve successful matching, the descriptors must be both repeatable and reliable~\cite{revaud2019r2d2}. However, challenges such as textureless environments, changes in illumination, and varying viewpoints make it difficult to generate unique and discriminative descriptors~\cite{zambanini2013local}.

\begin{figure}[t]
	\begin{center}
    \captionsetup{font=footnotesize}
    \begin{subfigure}[b]{0.47\textwidth}
        \includegraphics[width=1.0\textwidth]{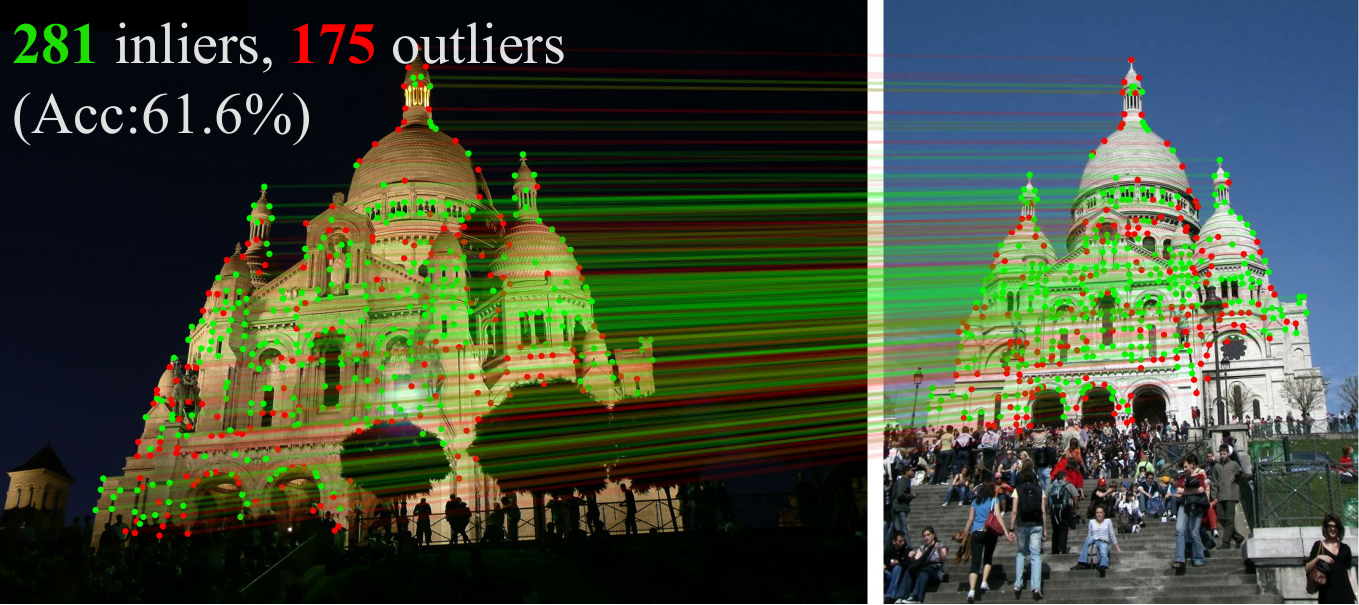}
        \caption{LightGlue~\cite{lindenberger2023lightglue}}
        \vspace{0.1cm}
    \end{subfigure}
    \begin{subfigure}[b]{0.47\textwidth}
        \includegraphics[width=1.0\textwidth]{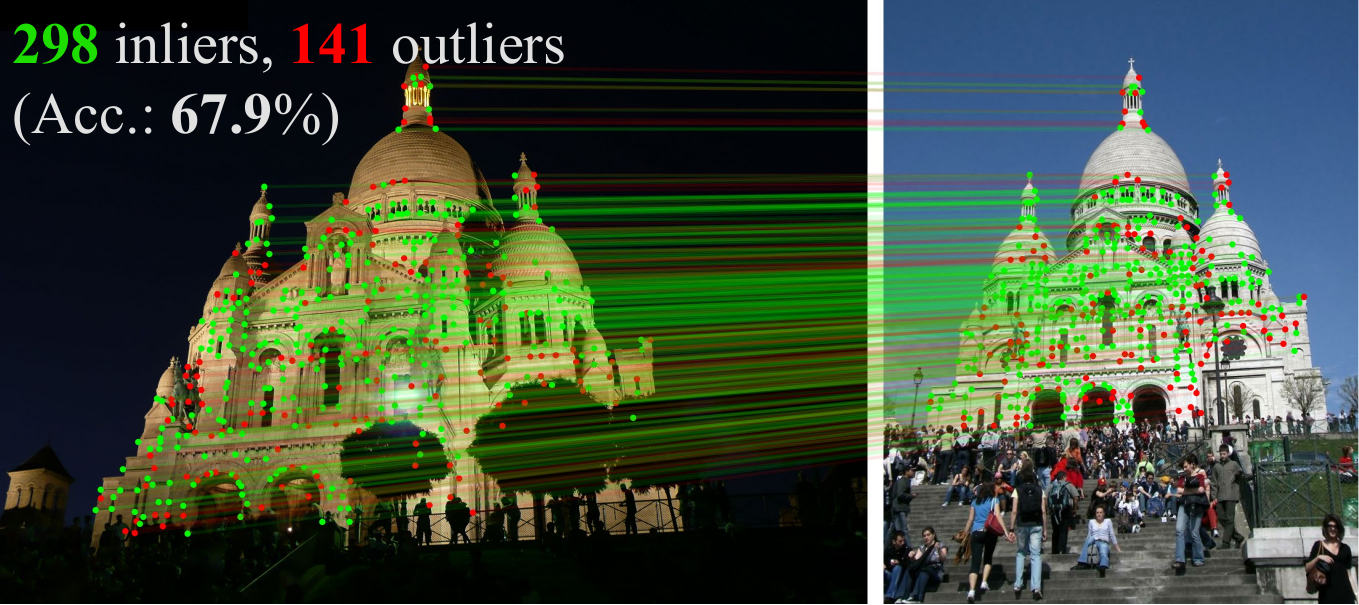}
        \caption{MambaGlue (Ours)}
    \end{subfigure}
    \caption{\protect\label{Visualization} Qualitative comparison of matching performance between LightGlue~\cite{lindenberger2023lightglue} and our proposed method called \textit{MambaGlue} on outdoor visual localization, given exactly the same keypoints and initial descriptors provided by SuperPoint~\cite{detone2018superpoint} under the same threshold parameters. Note that our MambaGlue demonstrates more robust matching performance even under challenging conditions, such as illumination changes, increasing the inlier ratio within the final correspondences.}
	\end{center}
 % \vspace{-0.7cm}
\end{figure}

To overcome the shortcomings of imperfect feature descriptors, researchers have studied various deep learning-based methods. In recent years, Transformers~\cite{vaswani2017attention} have become the de facto standard architecture in vision applications~\cite{caron2021emerging,dosovitskiy2020image,jaegle2021perceiver}, including feature matching. One of them is LoFTR~\cite{sun2021loftr}, which is a detector-free dense local feature matching model. It demonstrates better accuracy than the previous models~\cite{rocco2020efficient,li2020dual} by using Transformers in a coarse-to-fine manner. However, it is slow for applications that need low latency, like SLAM. Alternatively, sparse feature-based matching methods like SuperGlue~\cite{sarlin2020superglue} and LightGlue~\cite{lindenberger2023lightglue} have also been suggested. They also utilize the Transformer-based architecture~\cite{vaswani2017attention} for learning to match image pairs and demonstrate robust performance in feature matching in indoor and outdoor environments~\cite{sarlin2019coarse,sarlin2022lamar,sarlin2021back,sattler2018benchmarking}, also meeting a balance between speed and accuracy. However, the performance of the Transformer-based models still comes at a non-negligible amount of computing sources and training difficulty.
%Furthermore, to address the potential limitations of Transformer-based models still demanding a non-negligible amount of computing sources and training difficulty, Lindenberger~\textit{et al.} introduced LightGlue~\cite{lindenberger2023lightglue}, which improves upon SuperGlue by implementing simple yet effective structural changes for better efficiency and ease of training.

%However, the robust performance of SuperGlue~\cite{sarlin2020superglue} comes at the cost of high computational expense and training difficulty as with other Transformer-based models, requiring many computing resources that are inaccessible to most practitioners. To address potential limitations of SuperGlue, Lindenberger~\textit{et al.} introduced LightGlue~\cite{lindenberger2023lightglue}, which improves upon SuperGlue by implementing simple yet effective structural changes for better efficiency and ease of training.

In the meantime, Mamba~\cite{gu2023mamba}, an architecture recognized for its efficiency in handling sequential data, has been introduced recently. Because it can selectively focus on sequential input tokens, Mamba has been applied to language~\cite{lieber2024jamba} and also to vision~\cite{hatamizadeh2024mambavision,zhu2024vision,shi2024multi} tasks with prominent performance and fast speed in both training and inference.

In this paper, we propose a Mamba-based local feature matching model called \textit{MambaGlue}, which is a hybrid way of combining the Mamba architecture with the Transformer architecture. MambaGlue improves the performance of each layer that composes the overall model by leveraging Mamba's ability to selectively focus on input. Furthermore, we propose a network that predicts how much the estimated correspondences of the current layer are reliable.
By doing so, this module allows our MambaGlue to better determine whether to halt the iteration and ultimately reduce unnecessary computational costs. 
Our novel approach achieves significant improvement in accuracy with low latency by accurately adapting to the difficulty of feature matching for a given image pair.

\begin{figure*}[t]
	\begin{center}
    \captionsetup{font=footnotesize}
	\includegraphics[width=0.95\linewidth]{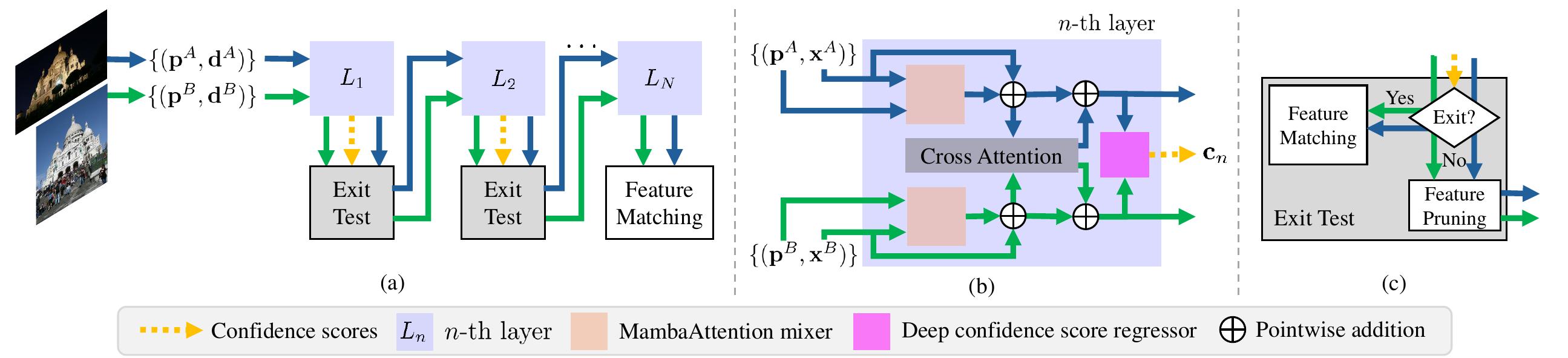}
	\caption{\protect\label{Network} (a)~Overview of the proposed feature matching pipeline called \textit{MambaGlue}. Pair sets of local feature points and their descriptors ($\mathbf{p}^I, \mathbf{d}^I$), where $I\in\{A,B\}$, pass through layers sequentially from $L_1$ to $L_N$, with an exit test at the end of each layer except the last layer. (b)~Description of the $n$-th layer in the pipeline, which mainly consists of a succession of a MambaAttention mixer, a cross-attention, and a deep confidence score regressor. Each layer augments the states $\mathbf{x}^A$ and $\mathbf{x}^B$, which are initialized by the local visual descriptors $\mathbf{d}^A$ and $\mathbf{d}^B$, respectively, i.e. $\mathbf{x}^A \leftarrow \mathbf{d}^A$ and $\mathbf{x}^B \leftarrow \mathbf{d}^B$, with global context as they pass through a \textit{MambaAttention mixer} and a cross-attention. At the end of the $L_n$ layer, where $n\in\{1,\ldots,N-1\}$, a deep confidence score regressor outputs the confidence scores set $\mathbf{c}_n$ to predict whether the current $n$-th matching prediction is sufficiently reliable. (c)~Diagram of the exit test. At the end of every layer, it decides whether to halt the process based on the confidence score. If enough number of features are confident for matching, \textit{MambaGlue} stops the iteration and performs feature matching; otherwise, the iteration proceeds after pruning potentially unreliable features.}
    \end{center}
    \vspace{-0.6cm}
\end{figure*}

%%%%%%%%%%%%%%%%%%%
%% MAIN CONTRIBUTION & WHAT FOLLOWS FROM THAT
% Explain your contribution in one paragraph. This is a very important paragraph. 
% Always start that paragraph with: ``The main contribution of this paper is''

% The main contribution of this paper is a \dots  
The main contributions of this paper are:
\begin{itemize}
	\item {To improve the performance of each layer, we propose a novel block called \textit{MambaAttention mixer} by utilizing the Mamba architecture, which is able to focus on input tokens selectively, with the attention architecture.} 
    \item {Furthermore, we propose a network called \textit{deep confidence score regressor} to predict the confidence scores indicating how much a feature point is reliably matchable.}
	\item As a result, our method achieves superior performance over the state-of-the-art methods with low latency.
	\item  In particular, it is remarkable that MambaGlue, a simple hybrid scheme of Mamba and Transformer, outperforms the state-of-the-art sparse feature matching methods.   
\end{itemize}

%%%%%%%%%%%%%%%%%%%
%% OUR KEY CLAIMS (can be merged with the main contribution above if desired)
% Explicitly(!) state your claims in one (short) paragraph and make
% sure you pick them up again in the experiments and support every claim.

% In sum, we make three key claims:
% Our approach is able to
% %
% (i) \dots;
% %
% (ii) \dots;
% %
% (iii) \dots.
% %
% These claims are backed up by the paper and our experimental evaluation.

%%%%%%%%%%%%%%%%%%%%%%%%%%%%%%%%%%%%%%%%%%%%%%%%%%%%%%%%%%%%%%%%%%%%%%%%%%%%%%%%
\section{Related Work}
\label{sec:related}

% Discuss the main related work and cite around 15-25 papers in sum. 
% The related work section should be approx. 1 column long, assuming 
% a 6-page paper.  Structure the section in paragraphs, grouping the 
% papers, and describing the key approaches with 1-2 sentences. If 
% applicable, describe the key difference to your approach at the end 
% of each paragraph briefly. Avoid adding subsections, al least for a 
% conference paper.

% The approach by Lim~\etalcite{lim2021ral} aims at predicting \dots

\subsection{Local Feature Matching}

%Matching images typically rely on local features. 
While numerous researchers have proposed novel image matching pipelines~\cite{philbin2007object,sivic2003video,pautrat2023gluestick,edstedt2024roma}, we place emphasis on the local feature-based image matching owing to its simple and intuitive functionality. The procedure for matching involves ({\romannumeral 1}) detecting interest points and representing the points with descriptors~\cite{lowe2004distinctive,bay2006surf,rublee2011orb,detone2018superpoint,dusmanu2019d2,revaud2019r2d2,yi2016lift}, ({\romannumeral 2}) matching these to make correspondences, ({\romannumeral 3}) filtering incorrect correspondences with techniques like random sample consensus (RANSAC), and finally ({\romannumeral 4}) estimating a geometric transformation matrix between an image pair with the final correspondences. 

In the procedure above, it is particularly crucial to establish correct correspondences while minimizing the number of spurious correspondences~\cite{Lim22icra-Quatro,Lim24ijrr-Quatropp}. The classical matcher is the nearest neighbor search~\cite{muja2009fast} in descriptor space. After matching, some correspondences are still incorrect because of imperfect descriptors or inherent noises. They are usually filtered out using heuristic methods, like Lowe's ratio test~\cite{lowe2004distinctive} or inlier classifiers~\cite{yi2018learning,zhang2019learning} and by robustly fitting geometric models~\cite{cavalli2020handcrafted,fischler1981random}. However, these heuristic processes require domain knowledge for parameter tuning and can easily fail under challenging conditions. These limitations of the matching are largely addressed by deep learning nowadays.

\subsection{Vision Transformer~(ViT)}

The introduction of Vision Transformer (ViT)~\cite{dosovitskiy2020image} revolutionized vision tasks, leading to methods like SuperGlue~\cite{sarlin2020superglue}, which combined ViTs with optimal transport~\cite{peyre2019computational} for improved feature matching. It is the first learning-based matcher that is trained to simultaneously match local features and filter out outliers from image pairs. By learning strong priors about scene geometry and camera motion, it demonstrates robustness to extreme variations and performs well across various data domains. However, like early Transformers~\cite{beltagy2020longformer,keles2023computational}, SuperGlue faces challenges, including being difficult to train and having computational complexity that scales quadratically with the number of keypoints.

%To tackle these problems, Lindenberger et al. proposed LightGlue, a subsequent work of SuperGlue that makes its design more efficient. Instead of reducing the network's overall capacity, LightGlue dynamically adapts its size based on the matching difficulty. It achieves this efficiency by incorporating techniques such as early stopping, feature pruning, and simpler matching processes, improving performance without sacrificing robustness

To tackle these problems, Lindenberger~\textit{et al.} proposed LightGlue~\cite{lindenberger2023lightglue}, a subsequent work of SuperGlue, that makes its design more efficient. Instead of reducing the network's overall capacity~\cite{chen2021learning,shi2022clustergnn}, LightGlue dynamically adapts its size based on the matching difficulty. It achieves this efficiency by incorporating techniques like early stopping, feature pruning, and simpler matching processes, improving performance without sacrificing robustness.

%Conversely, dense matchers like LoFTR~\cite{sun2021loftr} and other learning-based direct approaches~\cite{chen2022aspanformer,wang2022matchformer,wang2024efficient} match dense points rather than sparse ones. This improves the robustness impressively but is generally much slower than indirect approaches~\cite{sarlin2020superglue,lindenberger2023lightglue} because it processes dense features. This limits the resolution of the input images or leads to huge latency on high-resolution images. 

However, adding more Transformer-based structures to enhance the performance of LightGlue~\cite{lindenberger2023lightglue} can introduce additional computational complexity. To overcome the potential limitations of the Transformers, Mamba~\cite{gu2023mamba}, which aims to focus selectively on sequential data with linear-time complexity and selective state space updates, has emerged.

\subsection{Mamba Architecture and Hybrid Models}

Since the introduction of Mamba~\cite{gu2023mamba}, numerous novel approaches~\cite{shi2024multi,yang2024vivim} have been proposed to leverage its capability to capture long-range and spatio-temporal dependencies for vision applications. Specifically, Zhu~\textit{et al.} proposed Vision Mamba~\cite{zhu2024vision} that uses a bidirectional state space model (SSM) with the same Mamba formulation to capture more global context and improve spatial understanding. 

However, bidirectional encoding increases the computational load, which contradicts the advantage of Mamba and can slow down training and inference times. In addition, combining information from multiple directions effectively is challenging as some global context may be lost in the process. So far, models using only SSM architecture with causal convolutions are neither as efficient nor as effective as Transformer-only models. 

To resolve the potential limitations of the Mamba-only architectures, hybrid models~\cite{lieber2024jamba} utilizing Mamba-based architectures and Transformer-based architectures at the same time have emerged. Hatamizadeh and Kautz introduced MambaVision~\cite{hatamizadeh2024mambavision}, which is one of the hybrid methods. MambaVision uses a single forward pass with a redesigned Mamba block that can capture both short and long-range information and shows superior performance in terms of ImageNet top-1 throughput. It stacks its redesigned Mamba blocks and self-attention blocks with multi-layer perceptron (MLP) between blocks. Although adding MLP between blocks allows the network to extract richer high-dimensional features and then propagate them to the next layer, it is computationally expensive. Therefore, finding a way to utilize Mamba blocks with self-attention blocks with fewer resources is useful.

In this paper, we propose a novel parallel combination of Mamba and self-attention architectures for local feature matching. Unlike MambaVision that stacks Mamba and self-attention with MLP between them, our method connects them in parallel without MLP between them, resulting in more accurate performance with low latency.

%%%%%%%%%%%%%%%%%%%%%%%%%%%%%%%%%%%%%%%%%%%%%%%%%%%%%%%%%%%%%%%%%%%%%%%%%%%%%%%%
\section{The MambaGlue Architecture}
\label{sec:main}

%% Describe your approach. It is okay to divide the main section
%%  into a few subsections (e.g., 2-4 subsections).

% \subsection{How to Reference Materials}\label{subsec:test}

% References are already set as macro commands. Please refer to \texttt{URL-latex.tex}. All things are set! Just use \texttt{\textbackslash figref}, \texttt{\textbackslash secref}, \texttt{\textbackslash eqref}, and \texttt{\textbackslash tabref} as follows:

% \figref{fig:motivation}

% \secref{subsec:test}

% \subsection{Reference Test}\label{subsec:reference}

% \cite{lim2021ral}

% \subsubsection{Problem formulation}

\newcommand{\arbidx}{q}

The overall framework of the proposed feature matching method is shown in Fig.~\ref{Network}. Our MambaGlue mainly consists of a stacked layer pipeline with $N$ identical layers. The input to the system consists of two sets of local features from images $A$ and $B$. We denote feature sets of $A$ and $B$ as $\mathcal{F}_A$ and $\mathcal{F}_B$, respectively, formulated as follows:

\begin{equation}
    \label{feature-set}
    \begin{aligned}
    \mathcal{F}_A &= \{(\mathbf{p}_i^A, \mathbf{d}_i^A)\}^{N_A}_{i=1}, \\
    \mathcal{F}_B &= \{(\mathbf{p}_j^B, \mathbf{d}_j^B)\}^{N_B}_{j=1}.
    \end{aligned}
\end{equation}
Here, $i$ indicates an index of $\mathcal{A}$, which is the index set of $\mathcal{F}_A$, and $j$ indicates an index of $\mathcal{B}$, which is the index set of $\mathcal{F}_B$. $N_A$ and $N_B$ are the numbers of features on images $A$ and $B$, respectively, i.e. $|\mathcal{A}|=N_A$ and $|\mathcal{B}|=N_B$. For simplicity, an arbitrary $\arbidx$-th feature point and $d$-dimensional descriptor in $\mathcal{F}_A$ or $\mathcal{F}_B$ are denoted as $\mathbf{p}_\arbidx^I$ and $\mathbf{d}_\arbidx^I$, respectively, where $I\in\{A,B\}$.

The local features then pass through layers in the following order: a MambaAttention mixer, a cross-attention, and a deep confidence score regressor, as presented in Fig.~\ref{Network}(b), to enhance the expressiveness of the descriptors. At the end of the $n$-th layer, the deep confidence score regressor predicts set of the confidence scores $\mathbf{c}_n=\{c_q\,|\,q\in\mathcal{K}_n\}$, where $\mathcal{K}_n$ is the index set of all the features at the $n$-th iteration, i.e. $|\mathcal{K}_n|\leq|N_A|+|N_B|$. Then, the exit test determines whether to finish the iteration to reduce unnecessary computational cost. Otherwise, the features proceed to the feature pruning step, which rejects clearly unreliable features to enhance efficiency. If the system decides to halt the inference once enough correspondences are found, the iteration stops, and matching is performed to establish the correspondences. Consequently, the overall framework outputs a set of matches $\mathcal{M}=\{(i,j)\}\subset\mathcal{A}\times\mathcal{B}$. The pruning and matching steps, called the exit test, are the same as those in LightGlue~\cite{lindenberger2023lightglue}.

\begin{figure}[t!]
    \centering
    \captionsetup{font=footnotesize}
        \begin{subfigure}[b]{0.17\textwidth}
        \includegraphics[width=1.0\textwidth]{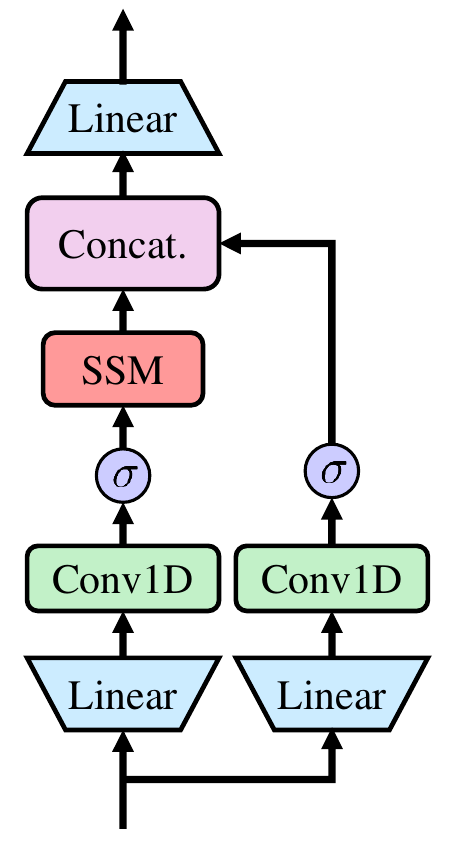}
        \caption{}
    \end{subfigure}
        \begin{subfigure}[b]{0.3\textwidth}
        \includegraphics[width=1.0\textwidth]{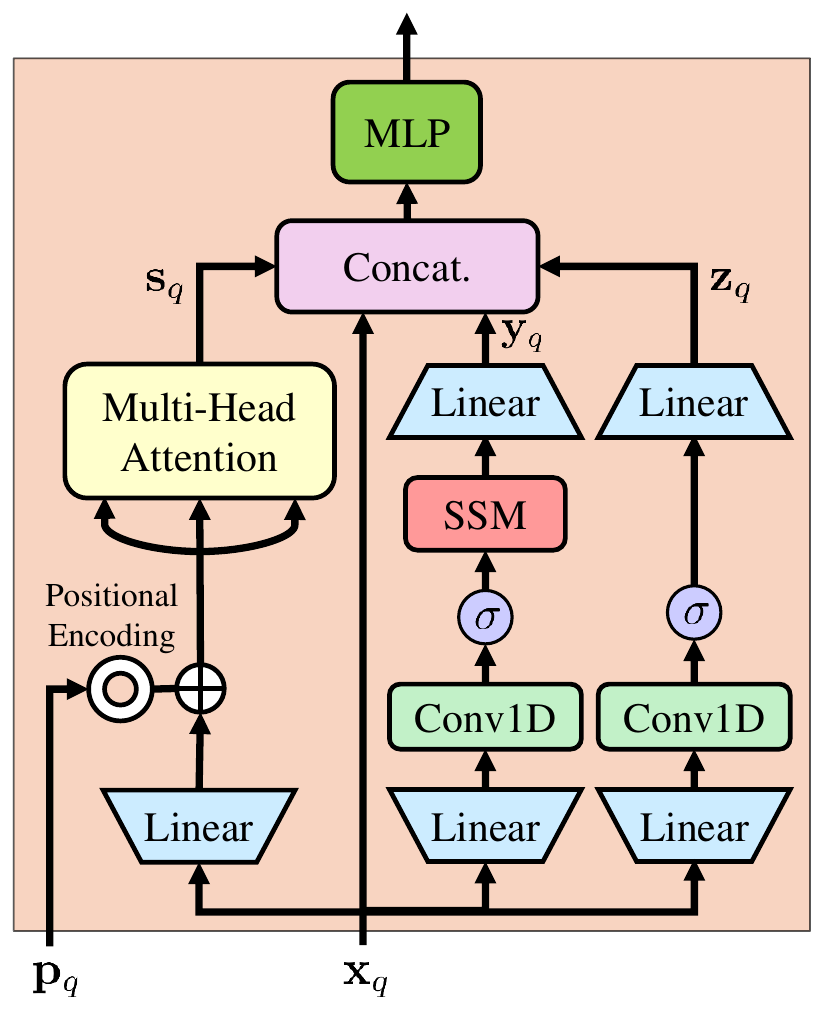}
        \caption{}
    \end{subfigure}
        \caption{\protect\label{MambaAttention Mixer}(a) The architecture of the MambaVision block~\cite{hatamizadeh2024mambavision}, which can only take an image as input and thus cannot be directly used for feature matching tasks, and (b) our proposed \textit{MambaAttention mixer} block, which takes feature points and states from descriptors as input. Our MambaAttention mixer mainly consists of three branches: (i)~a self-attention block with positional encoding for point input $\mathbf{p}_q$, (ii)~a direct connection of the input to preserve the original feature, and (iii)~a Mamba-based block, which is inspired by (a). Then, the features are concatenated at the end of the block to selectively and holistically provide the refined context for the next stage.}
    \vspace{-0.5cm}
\end{figure}

\newcommand{\state}{\mathbf{x}}
\newcommand{\destinationimage}{I}
\newcommand{\sourceimage}{S}

% arb: arbitrary
\newcommand{\arbstateforI}{\state_\arbidx^\destinationimage}
\newcommand{\arbdescforI}{\mathbf{d}_\arbidx^\destinationimage}

\newcommand{\messageComing}{\mathbf{m}_\arbidx^{I\mapsfrom{\sourceimage}}}
\newcommand{\messagebyselfmixer}{\mathbf{m}_\arbidx^{\destinationimage\leftarrow{\sourceimage}}}
\newcommand{\ithSelf}{\mathbf{s}_\arbidx}
\newcommand{\ithMambaY}{\mathbf{y}_\arbidx}
\newcommand{\ithMambaZ}{\mathbf{z}_\arbidx}

\newcommand{\projectionmatrix}{\mathbf{W}}

\newcommand{\mlplayer}{\mathrm{MLP}}
\newcommand{\concat}{[\cdot\,|\,\cdot]}
\newcommand{\linear}[2]{\mathrm{Linear}{#1 #2}}
\newcommand{\scan}{\mathrm{Scan}}
\newcommand{\activation}{\sigma}

\newcommand{\attentionscore}{a_{ik}^{\destinationimage\sourceimage}}

\newcommand{\rotaryencoding}{\mathbf{R}}
\newcommand{\sigmoidactivation}{\mathrm{Sigmoid}}
\newcommand{\confidenceclassifier}{c_i}

\newcommand{\simpliedconvlinfunction}{f(\state_\arbidx)}
\newcommand{\outputdim}{d}
\newcommand{\halfdim}{\frac{\outputdim}{2}}
\newcommand{\quarterdim}{\frac{\outputdim}{4}}

\subsection{MambaAttention Mixer} 

Inspired by MambaVision~\cite{hatamizadeh2024mambavision}~(see Fig.~\ref{MambaAttention Mixer}(a)), we begin by proposing a Mamba-based self-attention block, called MambaAttention mixer. As illustrated in Fig.~\ref{MambaAttention Mixer}(b), the MambaAttention mixer consists of a self-attention block, a direct connection of input, and a Mamba-based block. The combination of self-attention and Mamba allows for global and selective scanning of the input tokens.

Next, as shown in Fig.~\ref{Network}(b), the combination of the MambaAttention mixer block and the cross-attention block forms an essential part of each layer for our system. We assign the state $\arbstateforI\in\mathbb{R}^d$ to each $\arbidx$-th local feature in a target image $\destinationimage\in\{A,B\}$. Each state is initialized with the corresponding visual descriptor $\arbstateforI\leftarrow \arbdescforI$ and subsequently updated by the MambaAttention mixer block and the cross-attention block of each layer.

In both blocks, an MLP updates each state with a message $\messageComing$, which is the result of the aggregation from all states in a source image $S$ to a state in a target image $I$:
\begin{equation}
    \label{state-update}
    \arbstateforI\leftarrow \arbstateforI + \mlplayer([\arbstateforI\,|\,\messageComing]),
\end{equation}
where $\concat$ indicates the concatenation of two vectors. 
This is calculated simultaneously for all points in both images. In a MambaAttention mixer block, each image $I$ pulls information from points within the same image. In a cross-attention block, each image pulls information from the corresponding complement image.

For the sake of brevity, let us omit the superscript $I$. As shown in Fig.~\ref{MambaAttention Mixer}(b), the message by a MambaAttention mixer,~$\mathbf{m}_q$, is computed as the concatenation of outputs of Mamba-based paths $\ithMambaY$ and $\ithMambaZ$, and an output of self-attention path $\ithSelf$ as follows:
\vspace{-0.1cm}
\begin{equation}
    \label{message-by-MambaAttention Mixer}
    \mathbf{m}_\arbidx = [\ithSelf\,|\,\ithMambaY\,|\,\ithMambaZ],
\end{equation}
where $\ithSelf$ is computed as follows:
\begin{equation}
    \label{message-by-self-attention}
    \mathbf{s}_q = \sum_{j\in{\mathcal{I}}}\underset{k\in\mathcal{I}}{\text{Softmax}}(a_{qk})_j\mathbf{W}\mathbf{x}_j.
\end{equation}
Here, $\projectionmatrix$ is a projection matrix, $\mathcal{I}$ is the index set of $I$, $a_{qk}$ is an attention score defined as $a_{qk}=\mathbf{q}_q^\top\mathbf{R}(\mathbf{p}_k-\mathbf{p}_q)\mathbf{k}_k$, where $\mathbf{k}_i$ and $\mathbf{q}_i$ are the key and query vectors, respectively, generated by distinct linear transformations of an arbitrary state $\state_i$, % is decomposed into key and query vectors $\mathbf{k}_\arbidx$ and $\mathbf{q}_\arbidx$ through distinct linear transformations, 
and $\mathbf{R}(\cdot)\in\mathbb{R}^{d\times{d}}$ is a rotary encoding~\cite{su2024roformer} of the relative position between the points. Next, for simplicity, by denoting the encoding part, $\activation\left(\text{Conv}(\linear(\outputdim,\halfdim)(\state_\arbidx)))\right)$, as $f(\state_\arbidx)$, where $\linear(d_\text{in}, d_\text{out})(\cdot)$ denotes a linear layer with $d_\text{in}$ and $d_\text{out}$ as input and output embedding dimensions; $\text{Conv}(\cdot)$ is the convolutional layer and $\sigma$ is the Sigmoid Linear Unit (SiLU)~\cite{elfwing2018sigmoid} for activation, $\ithMambaY$ and $\ithMambaZ$ are defined as follows:
\vspace{-0.1cm}
\begin{equation}\label{message-by-mambavision}
    \begin{aligned}
    \ithMambaY &= \linear\Bigl(\halfdim, \outputdim\Bigr)\Bigl(\scan\bigl(\simpliedconvlinfunction\bigr)\Bigr), \\
    \ithMambaZ &= \linear\Bigl(\halfdim,\outputdim\Bigr)\Bigl(\simpliedconvlinfunction\Bigr),
    \end{aligned}
\end{equation}
where $\scan(\cdot)$ is the selective scan operation to efficiently focus on the most relevant segments of the input sequence~\cite{gu2023mamba}.

The $\messageComing$ by a cross-attention mechanism is computed as the weighted average of all states of image~$S$ as follows:
\begin{equation}
    \label{message-by-cross-attention}
\mathbf{m}_q^{I\mapsfrom{S}}=\sum_{j\in{\mathcal{S}}}\underset{k\in\mathcal{S}}{\text{Softmax}}(a_{qk}^{IS})_j\mathbf{W}\mathbf{x}_j^S,
\end{equation}
where $\mathcal{S}$ is the index set of $S$, and the attention score is defined as $a_{qk}^{IS}=\mathbf{k}_q^{I\top}\mathbf{k}_k^S$, where $\mathbf{k}_i$ is the key vector of an arbitrary state $\state_i$. Each point in $I$ attends to all points in the other image $S$. Thus, we need to compute the similarity only once for messages from both directions~\cite{hiller2024perceiving}.

\subsection{Deep Confidence Score Regressor}

The newly designed regressor, called the deep confidence score regressor, predicts a confidence score that indicates how confidently matching predictions are identical to ground truth matches for each feature point. Note that it is applied at the end of every $n$-th layer where $n\in\{1,\ldots,N-1\}$ (see Figs.~\ref{Network}(a) and~\ref{Network}(b)).

A combination of a sigmoid and only one linear layer is used to predict a confidence score in LightGlue~\cite{lindenberger2023lightglue}. However, only one linear computation layer is not enough to analyze the complex representation of each state that has gone through many steps of the neural network. We experimentally observed that our regressor network, even with deeper layers, is faster in both training and inference compared with using just a single linear layer. Additionally, it provides a better understanding of hierarchical and abstract meanings in the context. 

Formally, the definition of each $q$-th confidence score $c_q$ is as follows:
\begin{equation}\label{confidence-classifier}
    c_q=\text{Sigmoid}\Bigl(\text{MLP}(d\rightarrow\frac{d}{2}\rightarrow\frac{d}{4}\rightarrow{1})\left(\mathbf{x}_q\right)\Bigr),
\end{equation}
\noindent where $\text{MLP}(d_1\rightarrow d_2\rightarrow \ldots\rightarrow d_{\text{out}})(\cdot)$ denotes multiple MLP layers, where the dimension of the final output is $d_\text{out}$. Thus, (\ref{confidence-classifier}) indicates whether the state of $q$-th feature is reliably matchable or not.

\subsection{Exit Test for Early Stopping}

We adopt the exit test for efficient early stopping and for saving inference time, as proposed by Lindenberger~\textit{et~al.}~\cite{lindenberger2023lightglue}, allowing it to be applied when a user chooses to utilize it. Assuming that the $q$-th point in image $A$ or $B$ is deemed confident if $c_q > \lambda_n$, where $\lambda_n$ is a user-defined score, the exit test,~$\psi(\mathbf{c}_n)$, is performed at the end of every layer and is defined as follows: 
\begin{equation}\label{exit-formula}
    \psi(\mathbf{c}_n)=\begin{cases} 1 & \text{if } g(\mathbf{c}_n) > \alpha, \\
    0 & \text{otherwise},
    \end{cases}
\end{equation}
where $g(\mathbf{c}_n)=({1}/{|\mathcal{K}_n|})\underset{q=1}{\overset{|\mathcal{K}_n|}{\sum}}\llbracket c_q > \lambda_n \rrbracket$ and $\llbracket\cdot\rrbracket$ represents the Iverson operator.
%where $N_{T_n}=|\mathcal{I}_n|\leq|\mathcal{A}|+|\mathcal{B}|$, and $N_{T_n}\leq N_{T_{n-1}}$. 
That is, (\ref{exit-formula}) implies that we stop the iteration when a sufficient ratio~$\alpha$ of all points on an image pair is confident.

%We decay $\lambda_n$ throughout the layers based on the validation accuracy of each regressor because it is less confident in early layers.

\subsection{Loss Function}

We train MambaGlue in two stages, similar to the training procedure in LightGlue~\cite{lindenberger2023lightglue}. Initially, we train the network to predict correspondences without exit tests, followed by solely training the deep confidence score regressors. The second step does not impact the performance of each layer. 

The matching prediction matrix $\mathbf{P}$ is supervised using ground truth labels derived from two-view transformations, where points from $A$ are mapped to $B$ and vice versa based on relative pose and depth. Ground truth correspondences~$\mathcal{M}$ are point pairs with low projection errors in both images and consistent depth, while points in $\Tilde{\mathcal{A}}\subset\mathcal{A}$ and $\Tilde{\mathcal{B}}\subset\mathcal{B}$ are labeled unreliable if projection or depth errors are relatively large. The loss function $\mathcal{L}$ is designed to minimize the log-likelihood of the matches predicted at each layer:
\vspace{-0.1cm}
\begin{equation}\label{loss}
    \begin{aligned}
    \mathcal{L}=-\dfrac{1}{N}\underset{n=1}{\overset{N}{\sum}}\Bigl(
    & \dfrac{1}{|\mathcal{M}|}\sum_{(i,j)\in\mathcal{M}}\log{}^n\mathbf{P}_{ij} \\ 
    + \dfrac{1}{2|\Tilde{\mathcal{A}}|}\underset{i\in\Tilde{\mathcal{A}}}{\sum}\log(1-{}^n\sigma_i^A)
    & + \dfrac{1}{2|\Tilde{\mathcal{B}}|}\underset{j\in\Tilde{\mathcal{B}}}{\sum}\log(1-{}^n\sigma_j^B)\Bigr).
    \end{aligned}
\end{equation}
Here, $\mathbf{P}_{ij}=\sigma^A_i\sigma^B_j\underset{k\in\mathcal{A}}{\text{Softmax}}(\mathbf{S}_{kj})_i\underset{k\in\mathcal{A}}{\text{Softmax}}(\mathbf{S}_{ik})_j$, where a matchability score is defined as $\sigma_i=\text{Sigmoid}(\text{Linear}(\mathbf{x}_i))$, which encodes the likelihood of the $i$-th point to have a corresponding point, and a pairwise score matrix is defined as $\mathbf{S}_{ij}=\text{Linear}(\mathbf{x}^A_i)^\top\text{Linear}(\mathbf{x}^B_j)$, which encodes the affinity of each pair of points to be in correspondence.
This loss balances the contributions of positive and negative labels, ensuring accurate early predictions.

Next, we train the deep confidence score regressor. As described in (\ref{confidence-classifier}), we minimize binary cross-entropy~\cite{shannon1948mathematical} to make the matching predictions identical to the ground truth matches. Let ${}^{n}m_q^A\in\mathcal{B}$ represent the index of the point in $B$ matched to $q$-th point in $A$ at $n$-th layer. The ground truth label for each point is $\llbracket {}^{n}m_q^A={}^{N}m_q^A \rrbracket$. The same binary cross-entropy is applied for $B$.

\begin{figure}[t]
	\begin{center}
    \captionsetup{font=footnotesize}
    \begin{subfigure}[b]{0.23\textwidth}
        \includegraphics[width=1.0\textwidth]{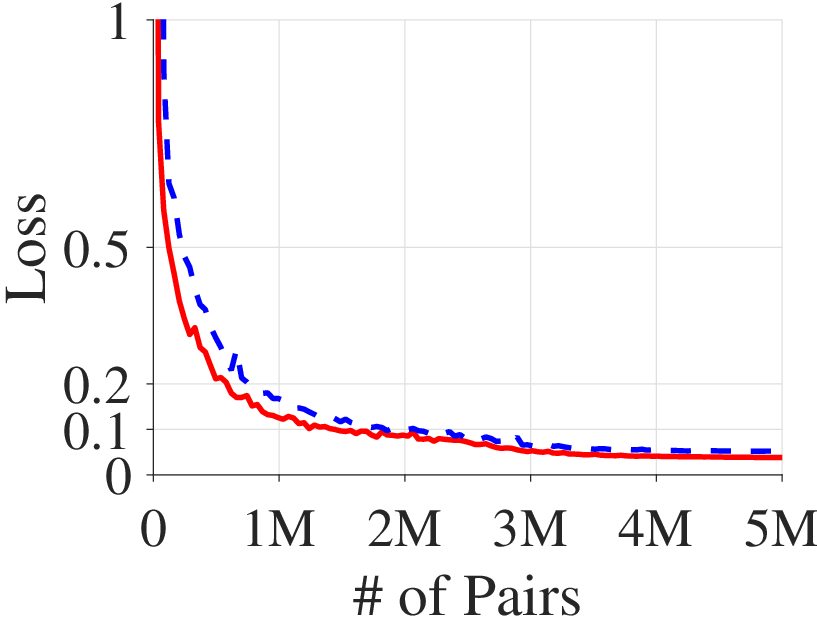}
        \caption{}
    \end{subfigure}
        \begin{subfigure}[b]{0.23\textwidth}
        \includegraphics[width=1.0\textwidth]{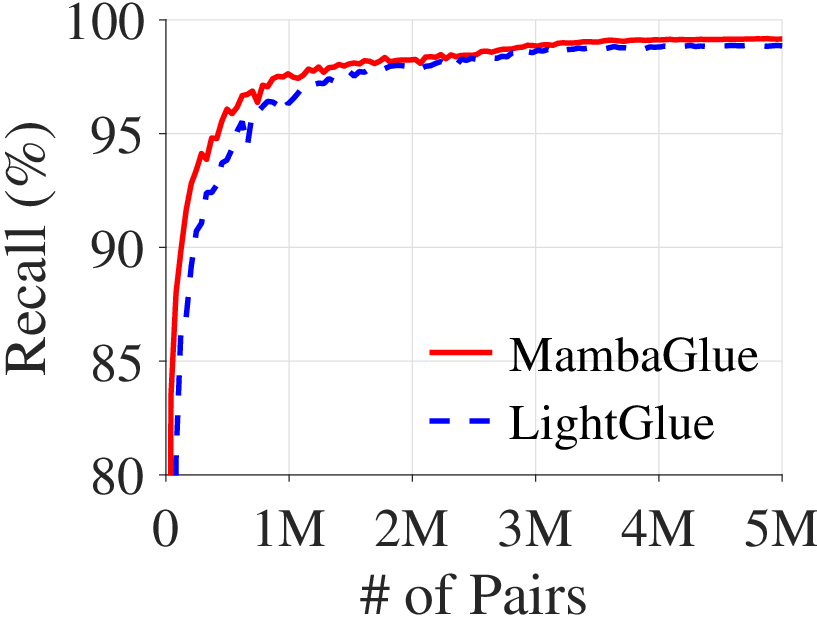}
        \caption{}
    \end{subfigure}
	\caption{\protect\label{Loss_Recall_graph} The loss and recall graph for the pre-training process of MambaGlue. After training on 5M image pairs (only 2 GPU-days), our MambaGlue achieves (a)~26.7\% lower loss at the final layer and (b)~0.3\% higher match recall than LightGlue.}
	\end{center}
 % \vspace{-0.7cm}
\end{figure}

\subsection{Comparison With LightGlue}

In summary, our MambaGlue is built on LightGlue but offers improved accuracy and efficiency. MambaGlue is more accurate at each layer and, thus, more accurate overall. By leveraging Mamba with self-attention, MambaGlue can selectively and globally process input, enhancing robustness beyond what is possible with transformer-based architectures alone. Additionally, at the end of each layer, the proposed deep confidence score regressor provides a hierarchical understanding of the state, resulting in contextually richer outputs compared with results using only a single linear layer. Despite all these improvements, the loss and recall graph shows that MambaGlue remains easy to train and even converges faster than LightGlue, as illustrated in Fig.~\ref{Loss_Recall_graph}.

%%% original
% \begin{table}[h!]
% 	\centering
%     \captionsetup{font=footnotesize}
%     \caption{Homography estimation on HPatches. MambaGlue produces better correspondences than sparse matchers, with the highest precision (P) @3px. The table shows the accurate homographies by MambaGlue when estimated by RANSAC or even a faster least-squares solver (DLT). MambaGlue is also competitive with dense matchers like LoFTR.}
% 	\begin{tabular}{lccccccc}
% 	\toprule
% 	\multicolumn{2}{c}{\multirow{2}{*}{features + matcher}} & \multirow{2}{*}{P} & \multicolumn{2}{c}{AUC - RANSAC} & \multicolumn{2}{c}{AUC - DLT} \\
% 	\cmidrule(lr){4-5} \cmidrule(lr){6-7}
% 	& & & @1px & @5px & @1px & @5px \\
% 	\midrule
% 	\multirow{3}{*}{\rotatebox{90}{\textbf{dense}}}
% 	& LoFTR & 92.7 & 41.5 & 78.8 & 38.5 & 70.6 \\
% 	& MatchFormer & 92.8 & 41.3 & 78.1 & 38.3 & 70.0 \\
% 	& ASpanFormer & 93.7 & 39.0 & 77.6 & 37.3 & 73.5 \\
% 	\midrule
% 	\multirow{5}{*}{\rotatebox{90}{\textbf{SuperPoint}}}
% 	& NN+mutual & 67.2 & 35.0 & 75.3 & 0.0 & 2.0 \\
% 	& SuperGlue& 87.4 & 38.3 & 79.3 & 33.8 & 76.7 \\
% 	& SGMNet & 83.0 & 38.6 & 79.0 & 31.7 & 76.0 \\
% 	& LightGlue & 88.9 & 38.3 & \textbf{79.6} & 35.9 & 78.6 \\
% 	& MambaGlue & \textbf{94.9} & \textbf{39.1} & \underline{79.4} & \textbf{37.5} & \textbf{78.9} \\
% 	\bottomrule
% 	\end{tabular}
% \end{table}

%%% reproduction
\begin{table}[t!]
    \captionsetup{font=footnotesize}
    \centering
    \caption{Comparison of the homography estimation on the HPatches dataset~\cite{balntas2017hpatches}. The precision with error threshold at 3\,px is denoted as~PR. The bold and the gray highlights denote the best for all cases and the best for feature-specific cases, respectively.}
    \label{table1-homography}
    \setlength{\tabcolsep}{4pt}
	\begin{tabular}{llccccccc}
	\toprule \midrule
        \multicolumn{2}{c}{\multirow{2}{*}[-0.3ex]{\raisebox{-0.3ex}{Extractor + Matcher}}} & \multirow{2}{*}[-0.3ex]{\raisebox{-0.3ex}{PR}} & \multicolumn{2}{c}{LO-RANSAC AUC} & \multicolumn{2}{c}{DLT AUC} \\
	\cmidrule(lr){4-5} \cmidrule(lr){6-7}
	& & & @1\,px & @5\,px & @1\,px & @5\,px \\
	\midrule
	\parbox[t]{2mm}{\multirow{3}{*}{\rotatebox[origin=c]{90}{Dense}}} & LoFTR~\cite{sun2021loftr} & 92.7 & \cellcolor{gray!30}\textbf{41.5} & \cellcolor{gray!30}78.8 & \cellcolor{gray!30}\textbf{38.5} & 70.6 \\
	& MatchFormer~\cite{wang2022matchformer} & 92.8 & 41.3 & 78.1 & 38.3 & 70.0 \\
	& ASpanFormer~\cite{chen2022aspanformer} & \cellcolor{gray!30}93.7 & 39.0 & 77.6 & 37.3 & \cellcolor{gray!30}73.5 \\
	\midrule
	\parbox[t]{2mm}{\multirow{5}{*}{\rotatebox[origin=c]{90}{SuperPoint}}}
	& NN+mutual~\cite{muja2009fast} & 67.2 & 34.6 & 74.5 & 0.4 & 3.4 \\
	& SuperGlue~\cite{sarlin2020superglue} & 87.4 & 37.1 & 78.7 & 32.1 & 75.8 \\
	& SGMNet~\cite{chen2021learning} & 83.0 & 38.6 & 79.0 & 31.7 & 76.0 \\
	& LightGlue~\cite{lindenberger2023lightglue} & 88.9 & 37.2 & 78.0 & 35.2 & 77.6 \\
	& MambaGlue (Ours) & \cellcolor{gray!30}\textbf{94.6} & \cellcolor{gray!30}39.0 & \cellcolor{gray!30}\textbf{79.3} & \cellcolor{gray!30}36.9 & \cellcolor{gray!30}\textbf{78.6} \\
	\midrule \bottomrule
	\end{tabular}	
 \vspace{-0.5cm}
\end{table}

%%%%%%%%%%%%%%%%%%%%%%%%%%%%%%%%%%%%%%%%%%%%%%%%%%%%%%%%%%%%%%%%%%%%%%%%%%%%%%%%
\section{Experimental Evaluation}
\label{sec:exp}

\subsection{Experimental Setups}

We evaluated MambaGlue on three visual tasks: homography estimation, relative pose estimation, and outdoor visual localization, by comparing it with Transformer-based sparse feature matching methods like SuperGlue~\cite{sarlin2020superglue}, SGMNet~\cite{chen2021learning}, and LightGlue~\cite{lindenberger2023lightglue} using their official pre-trained weights. The results of learning-based dense matchers~\cite{sun2021loftr,wang2022matchformer,chen2022aspanformer} are from LightGlue~\cite{lindenberger2023lightglue}.

%To be more concrete,
For homography estimation, we used the HPatches dataset~\cite{balntas2017hpatches}, which presents challenging conditions like illumination changes, occlusion, or viewpoint changes. In relative pose estimation, we used 1,500 image pairs from the MegaDepth-1500 dataset~\cite{li2018megadepth}, which includes outdoor scenes with structural and visual changes, with difficulty level adjusted by a visual overlap ratio. For outdoor visual localization, we employed the Aachen Day-Night benchmark~\cite{sattler2018benchmarking}, following the benchmark presented in Sarlin~\textit{et al.}~\cite{sarlin2019coarse}.

\subsection{Homography Estimation}

We assessed homography estimation accuracy using robust (LO-RANSAC with non-linear refinement~\cite{PoseLib}) and non-robust (the weighted DLT~\cite{hartley2003multiple}) estimators. LO-RANSAC leverages random sampling and local optimization to handle outliers effectively, whereas the DLT computes homography directly but is more prone to errors in the presence of noisy data. Evaluation metrics include the area under the curve (AUC) of the cumulative mean reprojection error with 1\,px and 5\,px, plus the precision at a 3\,px error threshold.

Table~\ref{table1-homography} shows that MambaGlue yielded correspondences with the highest precision. In particular, MambaGlue showed more accurate estimates than other sparse matchers, i.e.~approaches under the SuperPoint~\cite{detone2018superpoint} category in Table~\ref{table1-homography}, and was even competitive with matchers for dense features. At a coarse threshold of 5\,px, MambaGlue was even more accurate than LoFTR despite using sparse keypoints as input.

\begin{table}[t!]
        \captionsetup{font=footnotesize}
	\centering
        \caption{Relative pose estimation results on the MegaDepth1500 dataset~\cite{li2018megadepth}. Bold and gray highlights indicate the overall best and group-specific best performances, respectively, across three groups: ({\romannumeral 1}) dense features, ({\romannumeral 2}) superpoint + original model, and ({\romannumeral 3}) superpoint + exit test.}
	\label{table2-megadepth}
        % \scriptsize
	\setlength{\tabcolsep}{3pt}
	\begin{tabular}{llcccccccc}
	\toprule
	\multicolumn{2}{c}{\multirow{2}{*}[-0.3ex]{\raisebox{-0.3ex}{Extractor + Matcher}}} & \multicolumn{3}{c}{RANSAC AUC} & \multicolumn{3}{c}{LO-RANSAC AUC} & \multirow{2}{*}[-0.6ex]{\raisebox{-0.6ex}{\shortstack{Time\\(msec)}}} \\
	\cmidrule(lr){3-5} \cmidrule(lr){6-8}
	 & & 5° & 10° & 20° & 5° & 10° & 20° \\
	\midrule
	\parbox[t]{2mm}{\multirow{3}{*}{\rotatebox[origin=c]{90}{Dense}}}
	& LoFTR & 52.8 & 69.2 & 81.2 & 66.4 & 78.6 & 86.5 & 181.0 \\
	& MatchFormer & 53.3 & 69.7 & 81.8 & 66.5 & 78.9 & 87.5 & 388.0 \\
	& ASpanFormer & \cellcolor{gray!30}\textbf{55.3} & \cellcolor{gray!30}\textbf{71.5} & \cellcolor{gray!30}\textbf{83.1} & \cellcolor{gray!30}\textbf{69.4} & \cellcolor{gray!30}\textbf{81.1} & \cellcolor{gray!30}\textbf{88.9} & 369.0 \\
	\midrule
        \parbox[t]{2mm}{\multirow{7}{*}{\rotatebox[origin=c]{90}{SuperPoint}}}
	& NN+mutual & 30.0 & 45.7 & 59.3 & 49.9 & 62.6 & 72.2 & 5.7 \\
	& SuperGlue & 48.5 & 66.2 & 79.3 & 64.4 & 77.6 & 86.8 & 70.0 \\
	& SGMNet & 43.2 & 61.6 & 75.6 & 63.9 & 74.9 & 83.9 & 73.8 \\
	% & LightGlue & 49.8 & 66.8 & 79.9 & \cellcolor{gray!30}66.3 & \cellcolor{gray!30}78.8 & 87.5 & 44.2 \\
 %    & {\hspace{1em} \(\hookrightarrow\)} adaptive & 47.1 & 65.3 & 79.1 & 65.2 & 78.1 & 87.2 & 31.4 \\
	% & \textbf{MambaGlue} & \cellcolor{gray!30}50.1 & \cellcolor{gray!30}67.5 & \cellcolor{gray!30}80.3 & 65.8 & 78.7 & \cellcolor{gray!30}87.6 & 46.3 \\
 %    & {\hspace{1em} \(\hookrightarrow\)} \textbf{adaptive} & 49.4 & 66.8 & 79.9 & 65.4 & 78.3 & 87.3 & 35.3 \\
    & LightGlue & 49.8 & 66.8 & 79.9 & \cellcolor{gray!30}66.3 & \cellcolor{gray!30}78.8 & 87.5 & 44.2 \\    
	& Ours & \cellcolor{gray!30}50.1 & \cellcolor{gray!30}67.5 & \cellcolor{gray!30}80.3 & 65.8 & 78.7 & \cellcolor{gray!30}87.6 & 46.3 \\ \cmidrule(lr){2-9}
    & LightGlue + Exit test & 47.1 & 65.3 & 79.1 & 65.2 & 78.1 & 87.2 & 31.4 \\
    & Ours + Exit test & \cellcolor{gray!30}49.4 & \cellcolor{gray!30}66.8 & \cellcolor{gray!30}79.9 & \cellcolor{gray!30}65.4 & \cellcolor{gray!30}78.3 & \cellcolor{gray!30}87.3 & 33.1 \\
	\bottomrule
	\end{tabular}
 \vspace{-0.2cm}
\end{table}

%%%%%%%%%%%%%%%%%%%%%%%%
\subsection{Relative Pose Estimation}

For relative pose estimation, we calculated the essential matrix using RANSAC~\cite{fischler1981random} and LO-RANSAC with LM-refinement~\cite{PoseLib}, respectively. We calculated the pose error for pairs based on the maximum angular error in rotation and reported its AUC at $5^\circ$, $10^\circ$, and $20^\circ$.

\begin{table}[t!]
    \captionsetup{font=footnotesize}
	\centering
    \caption{Comparison of the outdoor visual localization on the Aachen Day-Night dataset~\cite{sattler2018benchmarking}. The bold and the underline denote the best and second-best performance, respectively.}
    \label{table3-aachen}
	\setlength{\tabcolsep}{3pt}
	\begin{tabular}{lcccc}
	\toprule
	\multicolumn{1}{c}{\multirow{2}{*}[-0.6ex]{\raisebox{-0.6ex}{\makecell{SuperPoint \\ + Matcher}}}} & Day & Night & \multirow{2}{*}[-0.6ex]{\raisebox{-0.6ex}{\makecell{Pairs per \\ second}}} \\
	\cmidrule(lr){2-3}
	 & \multicolumn{2}{c}{0.25\,m, 2° / 0.5\,m, 5° / 1.0\,m, 10°} \\
	\midrule
    NN+mutual & 84.8 / 90.3 / 93.8 & 65.3 / 72.4 / 85.7 & 174.3 \\
	SuperGlue & 88.7 / \textbf{95.5} / \textbf{98.7} & \underline{85.7} / \underline{92.9} / \textbf{100.0} & 6.5 \\
	SGMNet & 86.8 / 94.2 / \underline{97.7} & 83.7 / 91.8 / \underline{99.0} & 10.2 \\
	LightGlue & \underline{88.8} / 95.0 / 98.4 & \underline{85.7} / 91.8 / \underline{99.0} & 17.2 \\ % / 26.1
	MambaGlue (Ours) & \textbf{89.0} / \underline{95.3} / \textbf{98.7} & \textbf{86.7} / \textbf{93.9} / \textbf{100.0} & 16.7  \\ % / 25.7
	\bottomrule
	\end{tabular}
 \vspace{-0.4cm}
\end{table}

As shown in Table~\ref{table2-megadepth}, MambaGlue mostly showed promising performance compared with the state-of-the-art indirect approaches, such as SuperGlue, SGMNet, and LightGlue, given the same SuperPoint features with negligible additional processing time. Compared with LightGlue using the exit test, ours with the exit test showed a smaller performance degradation gap, while significantly boosting the inference speed. Considering the trade-off between accuracy and speed, we conclude that our MambaGlue achieves a balance between robustness and efficiency.

%%%%%%%%%%%%%%%%%%%%%%%%
\subsection{Outdoor Visual Localization}
Finally, for outdoor visual localization, we estimated camera poses with RANSAC and the perspective-n-point (PnP) solver. We reported the pose recall at multiple thresholds and the average throughput of the matching step during both mapping and localization.

As presented in Table~\ref{table3-aachen}, our MambaGlue demonstrated a substantial performance increase compared with other local feature matching methods, albeit with a slight trade-off in speed against the baseline pipeline~\cite{lindenberger2023lightglue}.

\subsection{Ablation Study}

\begin{figure}[t!]
    \centering
    \captionsetup{font=footnotesize}
        \begin{subfigure}[b]{0.151\textwidth}
        \includegraphics[width=1.0\textwidth]{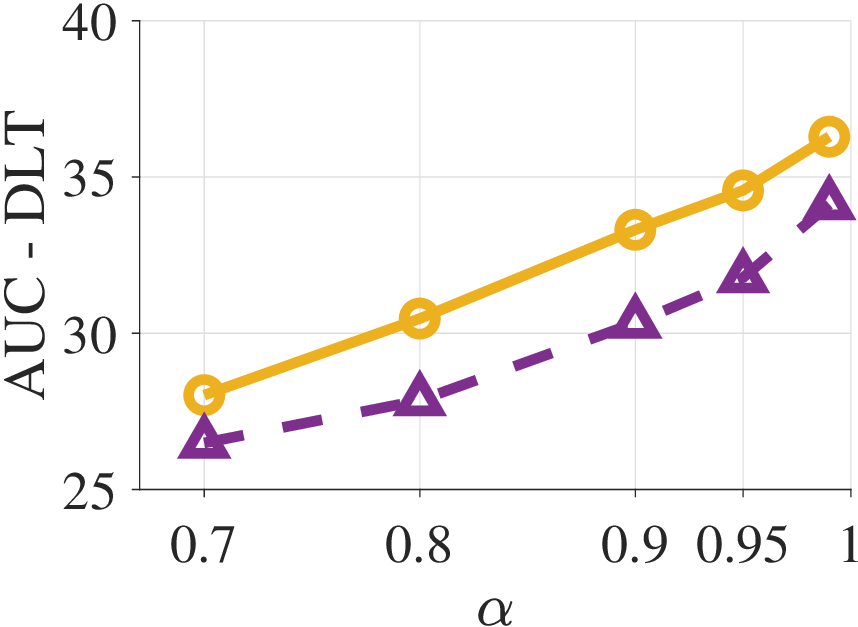}
        \includegraphics[width=1.0\textwidth]{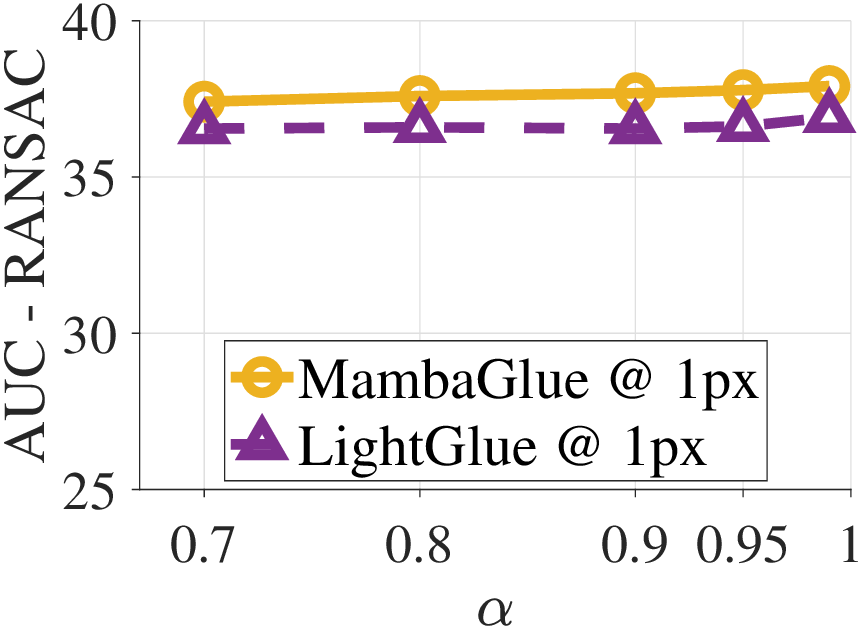}
        \caption{}
    \end{subfigure}
        \begin{subfigure}[b]{0.152\textwidth}
        \includegraphics[width=1.0\textwidth]{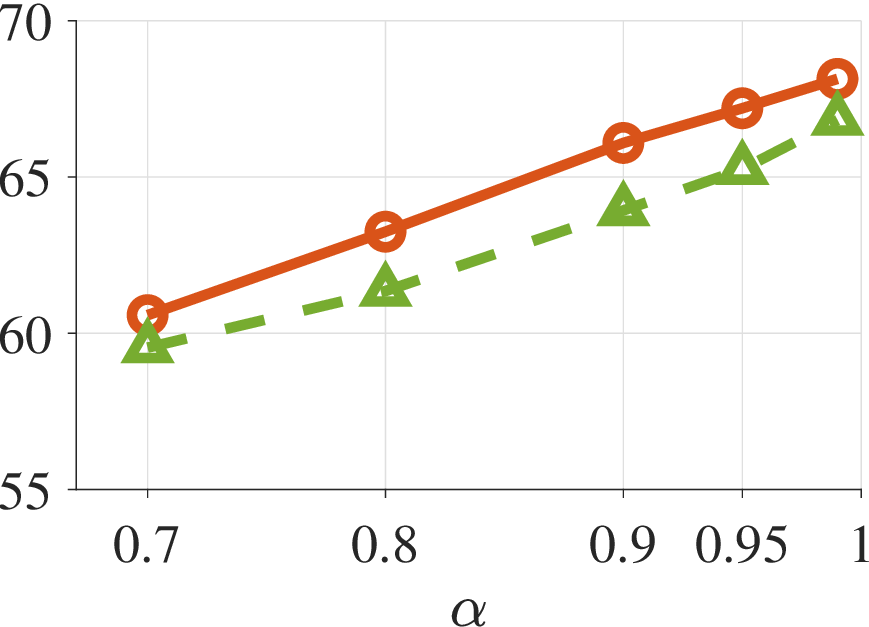}
        \includegraphics[width=1.0\textwidth]{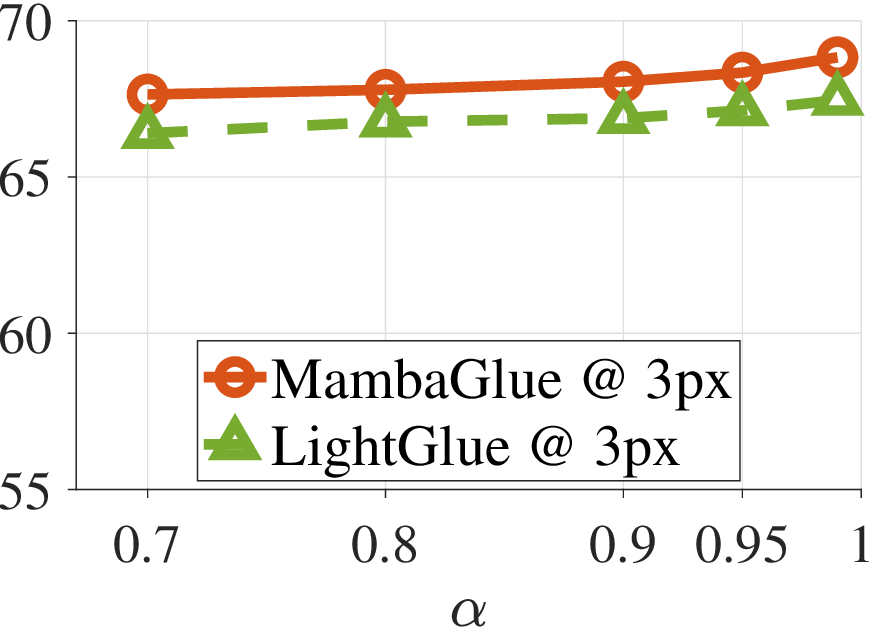}
        \caption{}
    \end{subfigure}
        \begin{subfigure}[b]{0.152\textwidth}
        \includegraphics[width=1.0\textwidth]{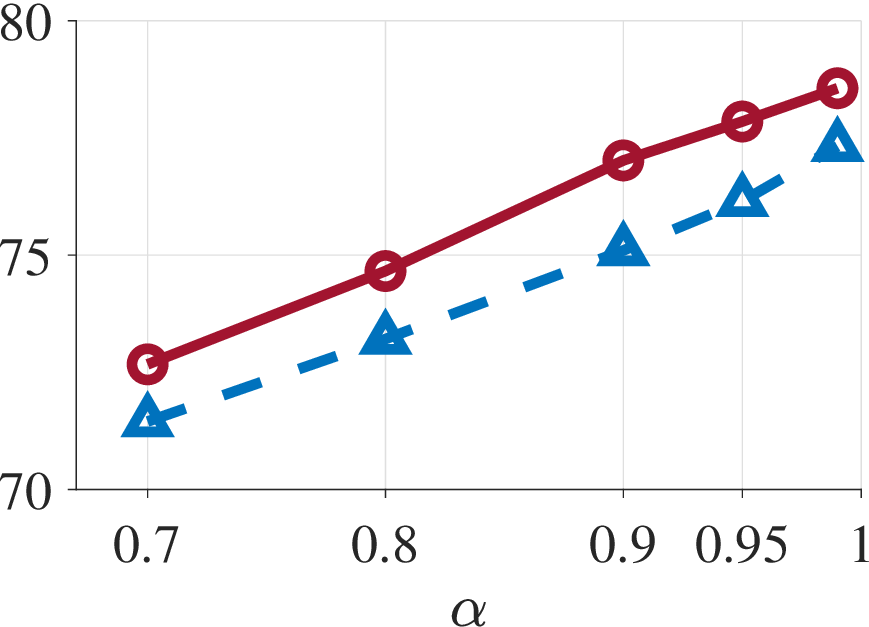}
        \includegraphics[width=1.0\textwidth]{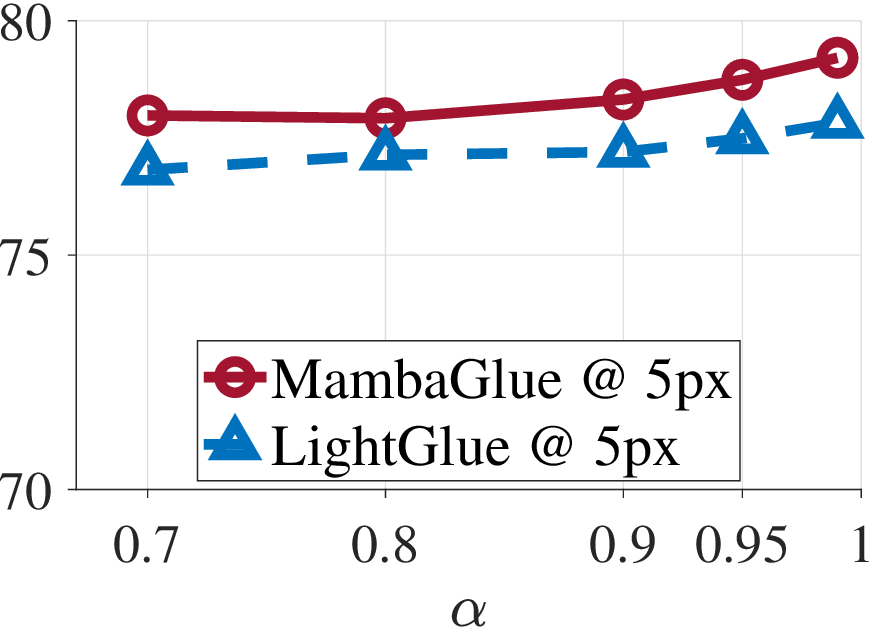}
        \caption{}
    \end{subfigure}   
    % \vskip\baselineskip
    % \includegraphics[width=0.5\linewidth]{pics/ablation_num_of_layers.eps}
        \caption{\protect\label{Ablation_Study}The AUC graph of reprojection error with varying exit thresholds $\alpha$ on the HPatches dataset~\cite{balntas2017hpatches} when using direct linear transformation~(DLT) and RANSAC with varying thresholds: (a)~1 pixel, (b)~3 pixels, (c)~5 pixels.}	
    \vspace{-0.1cm}
\end{figure}

% LightGlue의 Exit test랑 비교해보니 우리 것이 더 적절하게 빠져나오더라
We validated our model by comparing the exit test behavior between MambaGlue and LightGlue~\cite{lindenberger2023lightglue} using the homography dataset~\cite{balntas2017hpatches}. As shown in Fig.~\ref{Ablation_Study}, MambaGlue outperformed LightGlue in all scenarios under varying $\alpha$, which is the threshold in (\ref{exit-formula}). 

\begin{figure}[t]
	\begin{center}
    \captionsetup{font=footnotesize}
    \begin{subfigure}[b]{0.22\textwidth}
        \includegraphics[width=1.0\textwidth]{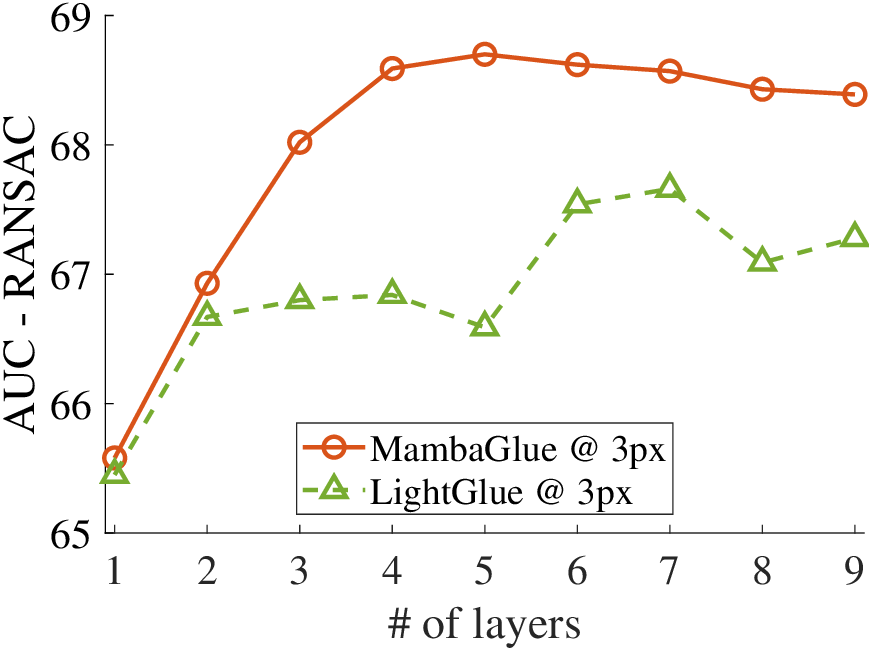}
        \caption{}
    \end{subfigure}
        \begin{subfigure}[b]{0.259\textwidth}
        \includegraphics[width=1.0\textwidth]{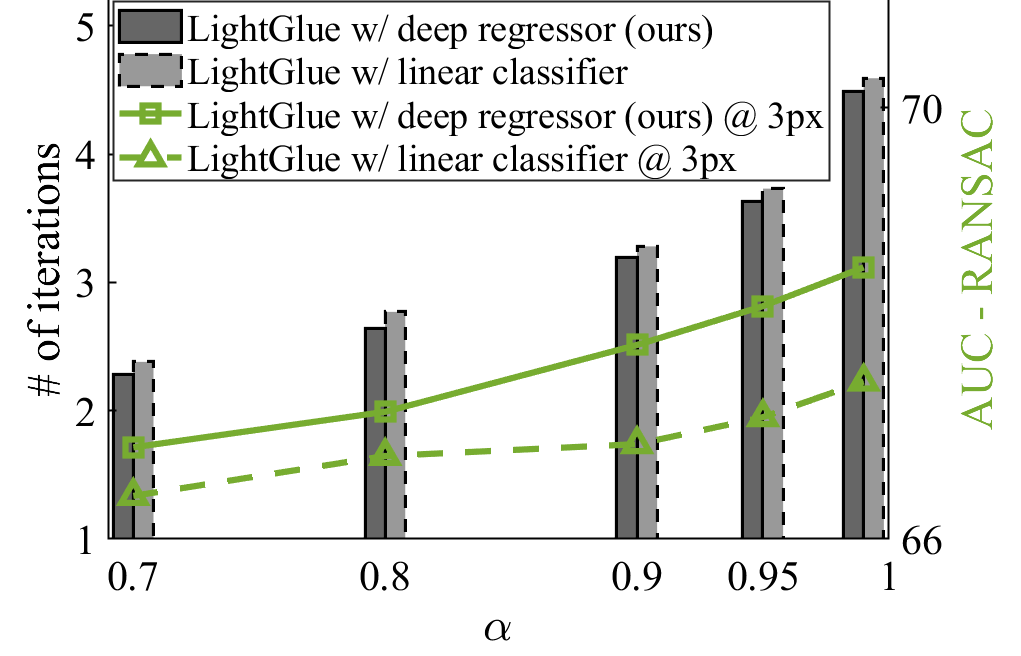}
        \caption{}
    \end{subfigure}
	\caption{\protect\label{Ablation_Study_n_regressor} (a) The AUC graph of reprojection error for models with varying number of layers without exit tests when using RANSAC with 3 pixels. (b) The graph for AUC of reprojection error and the number of iterations taken for the entire process with varying exit threshold $\alpha$ when using RANSAC with 3 pixels. The bar plot indicates the number of iterations, while the line plot shows AUC values. LightGlue with our deep confidence score regressor is referred to as LightGlue w/ deep regressor, while the original LightGlue is referred to as LightGlue w/ linear classifier.}
	\end{center}
 % \vspace{-0.5cm}
\end{figure}

When the number of layers in a model is limited without an exit test, as shown in Fig.~\ref{Ablation_Study_n_regressor}(a), MambaGlue outperformed LightGlue in terms of accuracy per layer, starting from the first layer, also showing more stable behavior as the number of layers increased. Next, as shown in Fig.~\ref{Ablation_Study_n_regressor}(b), our deep confidence score regressor inspects features with stricter criteria to stop at a more precise moment than LightGlue's confidence classifier, thus demonstrating improved performance with fewer iterations. Note that we also observed MambaGlue stopping the iterations at around the 5th iteration before the slope of the AUC graph in Fig.~\ref{Ablation_Study_n_regressor}(a) starts to decrease. 

Consequently, this experiment corroborates that the combination of the proposed modules efficiently improves the performance of feature matching, as presented in Fig.~\ref{Ablation_Study}.
%As shown in Fig.~\ref{num_of_layers}, "blahblahblah" showed that MambaGlue outperforms LightGlue in terms of accuracy per layer, with an average AUC increase of 1.86\% compared with 1.64\% for LightGlue.
%The value representing the difference in the ratio of AUC in Fig.~\ref{Ablation_Study}, divided by the difference in the number of iterations taken for the entire process in Fig.~\ref{num_of_layers}, showed that MambaGlue outperforms LightGlue in terms of accuracy per layer, with an average AUC increase of 1.86\% compared with 1.64\% for LightGlue.
%Interestingly, our deep confidence score classifier tends to require more iterations. This can be interpreted as inspecting features with stricter criteria in order to extract more reliable descriptors. For this reason, our proposed module significantly improves matching performance, as presented in Fig.~\ref{Ablation_Study}.

%%%%%%%%%%%%%%%%%%%%%%%%%%%%%%%%%%%%%%%%%%%%%%%%%%%%%%%%%%%%%%%%%%%%%%%%%%%%%%%%
\section{Conclusion}
\label{sec:conclusion}

In this paper, we have proposed a fast and robust matching method called \textit{MambaGlue}, which integrates Mamba and Transformer architectures for accurate local feature matching with low latency. In particular, we propose the MambaAttention mixer block to enhance the capability of self-attention and the deep confidence score regressor for predicting reliable feature matches. Our results demonstrate that MambaGlue strikes an optimal balance between accuracy and speed. 

Despite the successful improvement of our proposed method in terms of feature matching, our model's reliance on the Transformer architecture still demands a non-negligible amount of computational resources compared with Mamba architecture. In future works, we plan to make a Mamba-only model for more lightweight and faster feature matching.
% Our approach operates \dots  Our method exploits \dots
% This allows us to successfully \dots
% We implemented and evaluated our approach on different datasets
% and provided comparisons to other existing techniques and supported
% all claims made in this paper. The experiments suggest that \dots

%%%%%%%%%%%%%%%%%%%%%%%%%%%%%%%%%%%%%%%%%%%%%%%%%%%%%%%%%%%%%%%%%%%%%%%%%%%%%%%%
%% Future work: Use only if applicable -- but if so, use the following
%% sentence to start:
% Despite these encouraging results, there is further space for improvements. 

%%%%%%%%%%%%%%%%%%%%%%%%%%%%%%%%%%%%%%%%%%%%%%%%%%%%%%%%%%%%%%%%%%%%%%%%%%%%%%%%
% Only if applicable
%\section*{Acknowledgments}
%We thank XXX for fruitful discussions and for \dots

\bibliographystyle{URL-IEEEtrans}

% All new citations should go to new.bib. The file glorified.bib should go
% be the one from the ipb server. After paper or related work has been
% written merge the entries from new.bib to glorified.bib ON THE SERVER,
% replace the glorified.bib in this repository and empty the new.bib
\bibliography{URL-bib}

\end{document}